\documentclass{IEEEtran}
\usepackage{epsfig}
\usepackage{graphicx}
\usepackage{amsmath}
\usepackage{amssymb}

\usepackage[breaklinks=true,bookmarks=false]{hyperref}

\newfont{\mycrnotice}{ptmr8t at 7pt}
\newfont{\myconfname}{ptmri8t at 7pt}
\let\crnotice\mycrnotice%
\let\confname\myconfname%

\begin{document}



\title{Human Curation and Convnets: Powering Item-to-Item Recommendations on Pinterest}

\author{Dmitry Kislyuk$^{^1}$, Yuchen Liu$^{^{1,2}}$, David Liu$^{^1}$, Eric Tzeng$^{^{1,3}}$, Yushi Jing$^{^1}$ \\{$^{^1}$Visual Discovery \& Recommendations, Pinterest, Inc.}\\
$^{^2}$University of California, Los Angeles\\
$^{^3}$University of California, Berkeley\\
\{dkislyuk, yuchen, dliu, etzeng, jing\}@pinterest.com}
\maketitle


\begin{abstract}
This paper presents Pinterest Related Pins, an item-to-item recommendation system that combines collaborative filtering with content-based ranking. We demonstrate that signals derived from \textit{user curation}, the activity of users organizing content, are highly effective when used in conjunction with content-based ranking. This paper also demonstrates the effectiveness of visual features, such as image or object representations learned from convnets, in improving the user engagement rate of our item-to-item recommendation system. 


\end{abstract}




\section{Introduction}

Pinterest is an online catalog used to discover and save ideas. Hundreds of millions of users organize Pins around particular topics by saving them to boards. Each of the more than 50 billion Pins saved on Pinterest has an image, resulting a large-scale, hand-curated collection, with a rich set of metadata.

Most Pins' images are well annotated: when a person bookmarks an image on to a board, a \textit{Pin} is created around an image and a brief text description supplied by the user.  When the same Pin is subsequently saved to a new board by a different user, the original Pin gains additional metadata that the new user provides. Therefore, the data structure surrounding each Pin continues to get richer each time the Pin is re-saved. Furthermore, boards (i.e. collections of Pins) reveal relations \textit{between} Pins: if many users save these two Pins together, there is a high likelihood that another user may find them to be related as well. Such aggregated \textit{image co-occurrence} statistics are found to be useful for related content recommendation.


\begin{figure}
\centering \includegraphics[width=2.8 in]{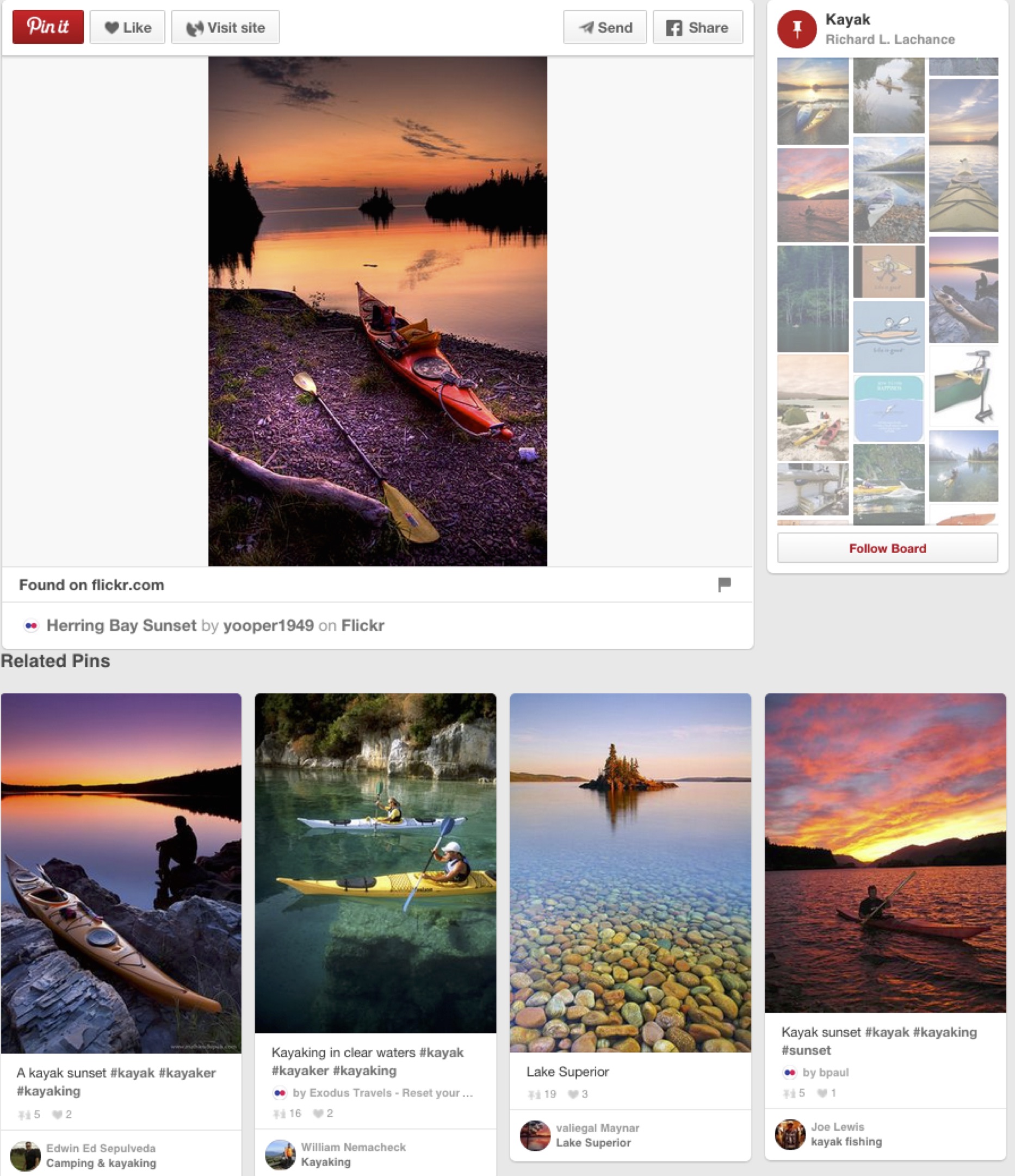}
\caption{Related Pins is a product that shows recommendations based on the Pinterest curation graph.}
\label{fig:relatedpins}
\end{figure}
\begin{figure}
\centering \includegraphics[width=2.8 in]{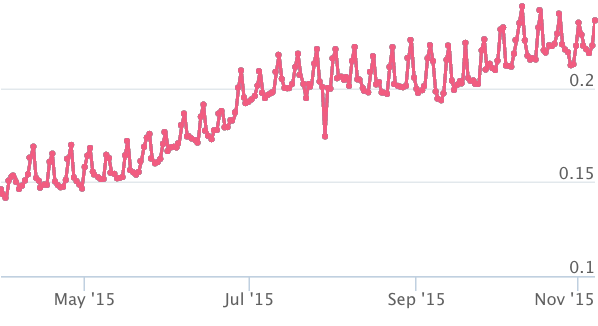}
\caption{Since May 2015, the percentage of users engaging with Related Pins recommendations, given they view a Pin closeup, has increased by 50\%}
\label{fig:related_pins_growth}
\end{figure}

This work explores how user curation signals can be used in conjunction with content-based features to improve recommendation systems. Specifically we introduce Related Pins, an \textit{item-to-item} content recommendation service triggered when a Pin closeup is shown to the user, and describe in detail our experiments using visual features (such as those obtained from convolutional neural networks), which are of particular interest since this system ultimately recommends visual content. As one of the most popular features on Pinterest~\footnote{More than 20\% of the page views originate from users clicking on recommended results.}, Related Pins is a recommendation system that combines collaborative filtering~\cite{Sarwar:2001} with content-based~\cite{Datta:2005} retrieval. Since May 2015, the user engagement metric~\footnote{Defined by the percentage of users clicking off-site or saving recommended content in one day given they see at least one Pin Recommendation.} on Pin recommendations has improved by more than 50\%.  Note that the improvement is the result of using both visual features and other metadata signals in the learning-to-rank framework---the scope of this paper is limited to the understanding of user curation and visual features in the context of recommendation systems.

This work makes two contributions: first, we demonstrate that ``Pinning,'' a form of user curation, provides valuable user signals for content recommendation. Specifically we present our use of \textit{image/board co-occurrences}, including the \textit{PinJoin} data structure used to derive this signal. Second, we demonstrate that combining collaborative filtering with content-based retrieval methods, such as applying a learning-to-rank framework~\cite{Joachims:2002} to a set of semantic and visual features associated with the candidate Pins, can significantly improve user engagement. In particular, our A/B experiments demonstrate that the use of recently developed visual features (when used in conjunction with other text and graph signals), such as those obtained from VGG~\cite{vggnet2014} and Faster R-CNN~\cite{frcnn15} yield significant gains in recommendation quality.


\section{Related Work}

Collaborative filtering~\cite{Sarwar:2001} using user-generated signals (e.g. co-views, co-clicks) is widely used in commercially deployed recommendation systems such as YouTube related videos~\cite{Baluja:2008,weston14awe,Bendersky:2014} and Amazon Related Items~\cite{linden}. This work investigates the use of \textit{user curation} signals derived from Pins' image/board co-occurrences, which are unique to Pinterest.

Visual features are widely used in both content-based recommendation systems and image search systems~\cite{Datta:2008,visualrank}, and the the learning-to-rank framework used in this paper from~\cite{Joachims:2002} has been widely used in industry~\cite{watson, frome07, jing12,Richardson:2006}. To our best knowledge this work contains the first published empirical results on how the latest convolutional neural network (CNN) based visual features (e.g. VGG~\cite{vggnet2014}) and large-scale object detection using Faster R-CNN\cite{frcnn15} can improve commercial recommendation systems.  Visual features are computed using a distributed process described in our previous work~\cite{Jing:2015}.

\section{User Curation Signals}

Content curation is the process of organizing and collecting content relevant to a particular topic of interest. Pinterest is a user-powered content curation service as content is collected and grouped into topic boards, creating a rich set of metadata associated with Pins' images. For example, during Pin creation, users typically provide a text description of the images as shown in Figure 3. Although any single instance of text description can be noisy, an aggregated collection reveals important annotations relevant to the Pin's image.  Furthermore, when a Pin is saved to a board, one can infer the categorical information of the Pin's image from the category the user selected for the board.

Formally, we denote the data structures associated with Pins and boards in the following way: each \textit{PinJoin} is a 3-tuple $p = \{u, P, A\}$, where $u$ is the image URL, $P$ is the collection of Pins generated for that image, and $A$ is the aggregation of text annotations or keywords (extracted from Board titles and descriptions). Each \textit{BoardJoin} is represented as a 2-tuple $b = \{t, P\}$, where t is the board title and P is a list of Pins. PinJoin is conceptually similar to Visual Synsets~\cite{TsaiICCV11} except in this case, all the images within a visual synset are exact duplicates of each other and the collection is curated manually. In practice, both of these structures contain additional metadata, such as category and topic information, which is used for deriving other features during re-ranking.

\begin{figure}
\centering \includegraphics[width=3.3 in]{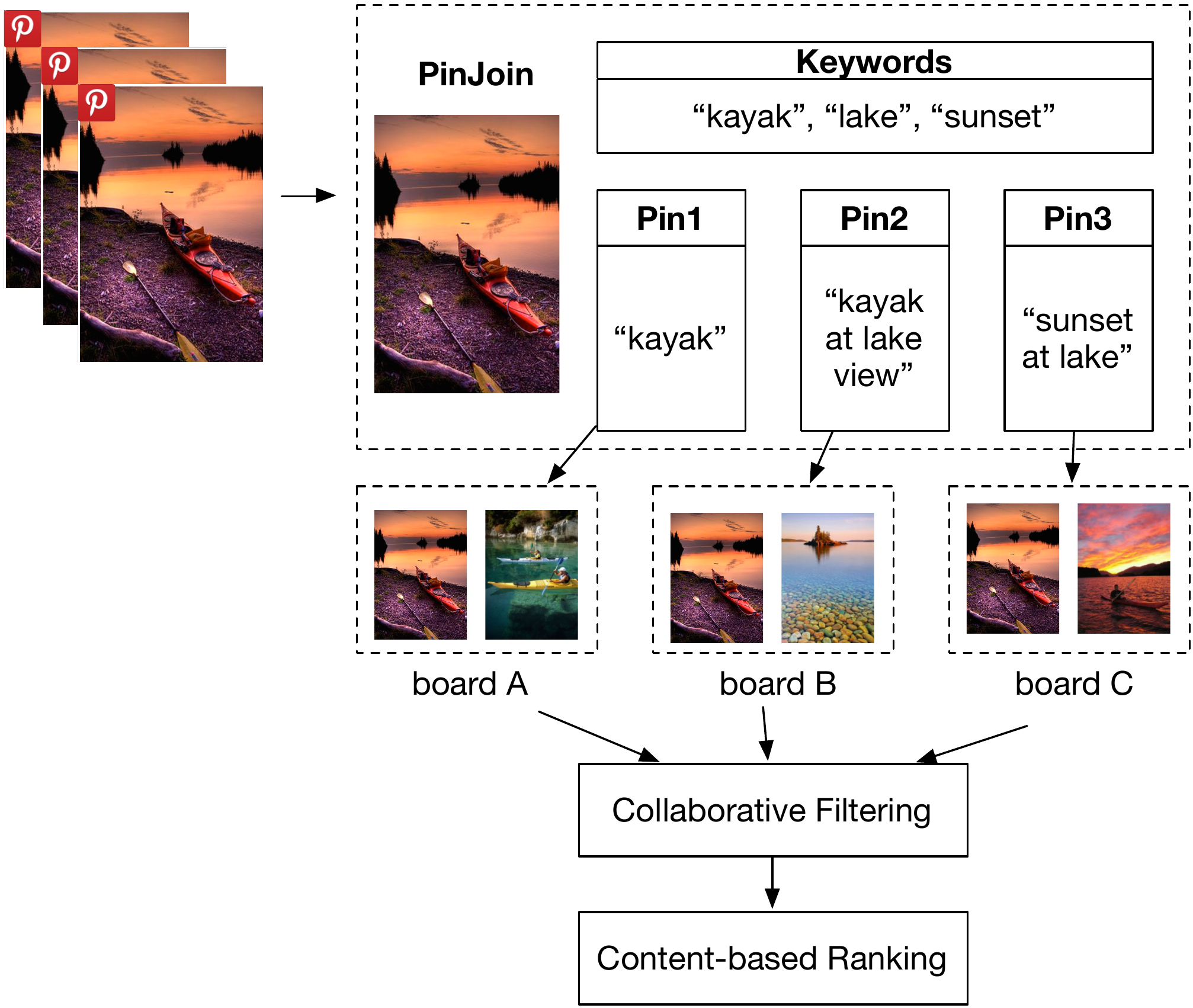}
\caption{Pins of the same image are aggregated together to form a \textit{PinJoin}, where rich metadata for the image could be generated based on information from different Pins by different Pinners.}
\label{fig:pinjoins}
\end{figure}

\begin{figure*}
\centering
\includegraphics[width=6.5in]{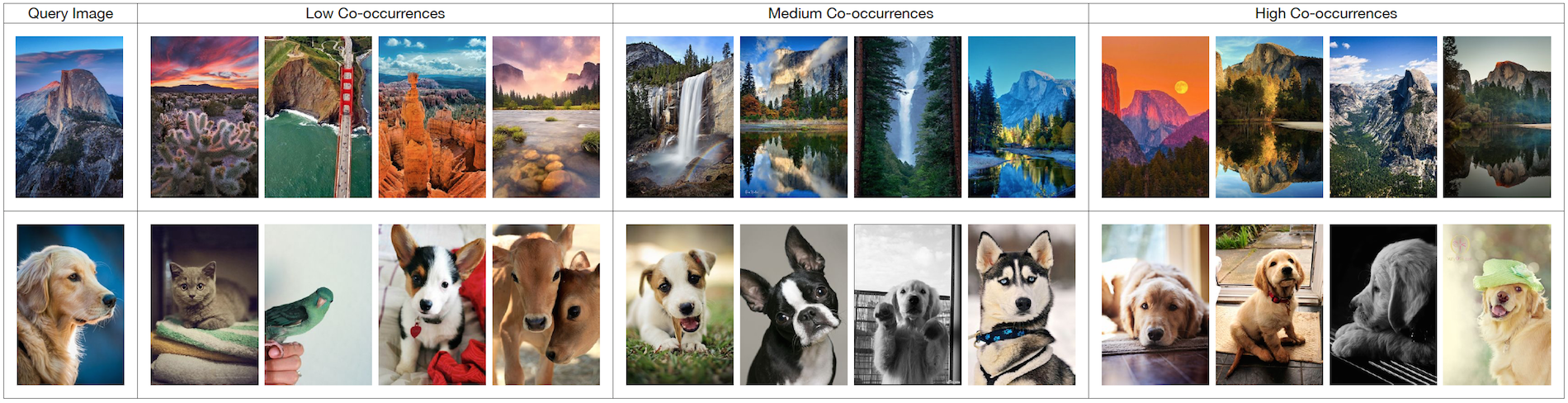}
\caption{Examples of candidates with low, medium and high board co-occurrences with the query image. The relevance of the candidate gradually increases with higher co-occurrences with the query image. \textit{Top example:} low: travel destinations, medium: Yosemite viewpoints, high: Half Dome. \textit{Bottom example:} low: animals, medium: dogs, high: golden retrievers.}
\label{fig:occur_examples}
\end{figure*}

User curation reveals relations \textit{among} images: we observed that images of Pins on the same board are semantically (and to some extent visually) related to each other. Therefore, if enough users save these two Pins together, there is a high likelihood that a new user may also find them to be related. This is helped by the fact that users on Pinterest actively curate content---our engaged users have an average of 24 boards. An example of image/board co-occurrences is shown in Figure~\ref{fig:occur_examples}. 

\section{Related Pin Recommendations}
This section presents the architecture that powers the Pinterest Related Pins Recommendation system.  The first step described relies on collaborative filtering over user curation signals (image co-occurrences on boards) to generate a candidate sets.  The second step uses content-based ranking approach to rank the candidates based on content signals such as visual features, textual signals, and categories signals derived from \textit{PinJoins}. 

\subsection{Candidate Generation from User Curation}

The first step of the pipeline is to generate a set of image candidates for each query image, which will serve as candidate sets for content-based re-ranking. We adopt a classic collaborative filtering approach to exploit the Pin/board co-occurrences as described in Section 3. For each Pin, we select up to 10,000 Pins with the highest number of shared boards. In practice, the candidate generation process is accomplished through a MapReduce job, which takes \textit{BoardJoin} $B = {b_1, b_2, ..., b_n}$ as input. The mapping stage outputs image pairs $(p_i, p_j)$ for all image pairs in each board $b$, and in the reduce stage all the related images are grouped by the same query image. For computational efficiency, we sample images based on the quality/popularity of the images. For Pins that do not generate enough candidates through board co-occurrence, we rely on a content-based retrieval service described in our previous paper~\cite{Jing:2015}.

\subsection{Content-based Ranking}

After generating a set of candidates, we re-rank with a set of both content pair-features (defined between a query and a candidate) and query-independent features. In addition to standard features such as text annotation match and topic vector similarity, we were particularly interested in the effectiveness of visual similarity features for our re-ranking step. Examples of visual features include the \textit{fc6} and \textit{fc8} activations of intermediate layers of deep convolutional neural networks (CNNs)~\cite{decaf} based on AlexNet~\cite{alexnet12} and VGG~\cite{Simonyan14c}. These features are binarized (\textit{fc6}) and sparsified (\textit{fc8}) for representation efficiency and compared using Hamming distance (\textit{fc6}) and cosine similarity (\textit{fc8}), respectively.  We use the open-source Caffe~\cite{jia2014caffe} framework to perform training and inference of our CNNs on multi-GPU machines. In this work, we also trained an object detection module using Faster R-CNN~\cite{frcnn15}, initially fine-tuned on a dataset containing the most common objects found on Pinterest, including home decor and fashion categories such as various furniture types, shoes, dresses, glasses, bags and more. 

To learn the weight vector for our linear model, we adopted the learning-to-rank approach from Joachims~\cite{Joachims:2002}. Given training data in the form of relative ranking triplets $(q^{(k)}, d_+^{(k)}, d_-^{(k)})$, where document $d_+^{(k)}$ is considered to be more relevant to query $q^{(k)}$ than document $d_-^{(k)}$, the RankSVM algorithm described in~\cite{Joachims:2002} approximates a weight vector $\vec w$ which maximizes the number of training examples satisfying $\vec{w} \Phi(q, d_+) > \vec{w} \Phi(q, d_-)$, where $\Phi(q, d)\in \mathbb{R}^m$ gives $m$ features of the document $d$ in the context of query $q$.

The relevance triplets we use in training are generated through user clicks and impression logs, normalized by position and device to account for position bias, using a clicks over expected clicks~\cite{coec} (COEC) model. For each query $q^{(k)}$ in our training set, given the set of observed results $D^{(k)}$, we generate the training triplet: $$q^{(k)},\:  argmax_{d \in D^{(k)}}\textsc{coec}(d),\: argmin_{d \in D^{(k)}}\textsc{coec}(d)$$ corresponding to the query, best engaged document, and worst engaged document. We also generate random negative examples:
$$q^{(k)},\: argmin_{d \in D^{(k)}}\textsc{coec}(d),\: d_{\textsc{random}}$$
on the intuition that even poorly engaged candidates generated through our board co-occurrence signal should still be more relevant than a random document from our corpus of Pins.

\section{Experiments}

\subsection{Analysis of User Curation Signals}

\begin{figure}
\centering \includegraphics[width=3.0in]{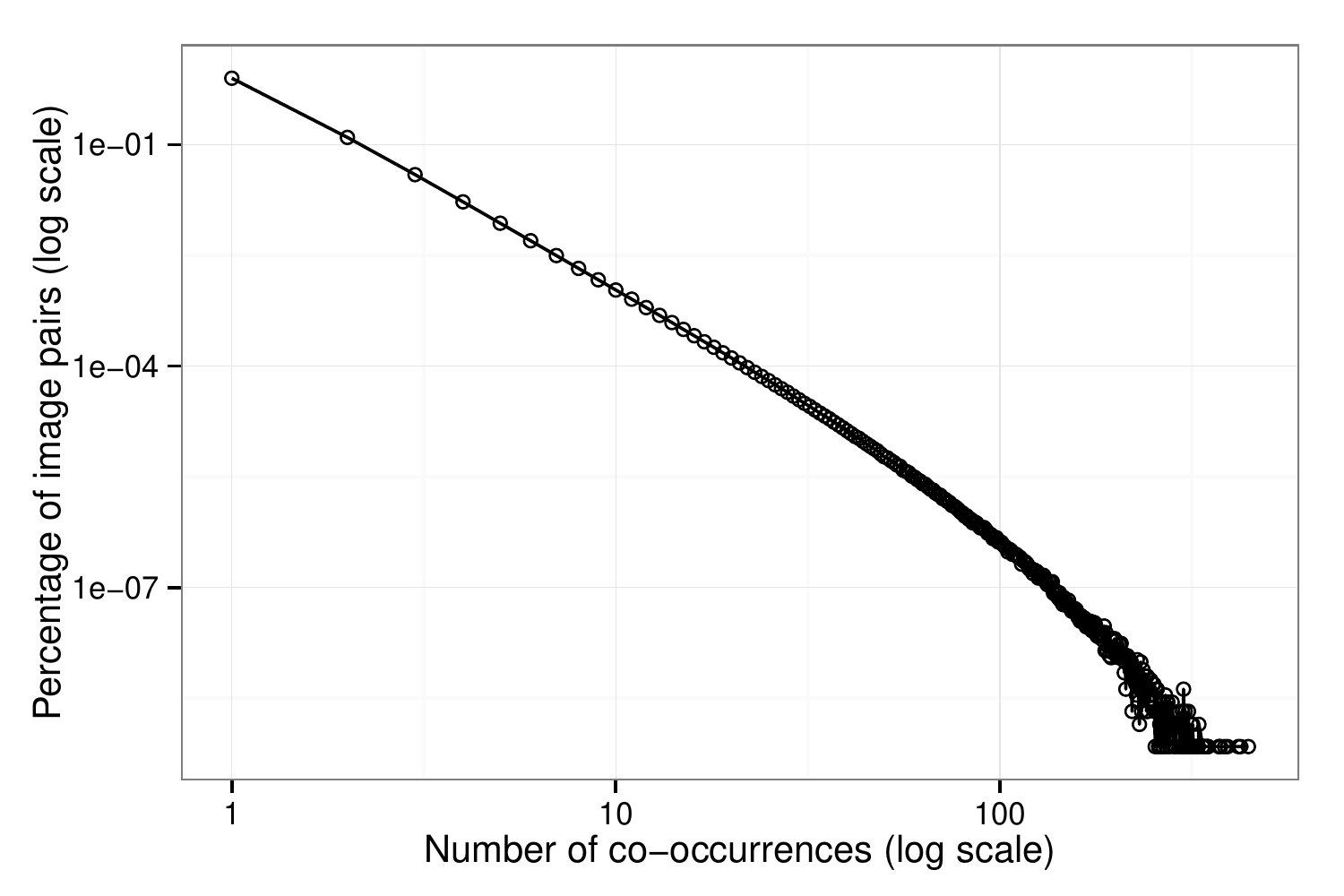}
\caption{Percentage of image pairs on Pinterest having different number of board co-occurrences: majority of pairs (80\%) co-occur on boards once.}
\label{fig:cooccurrence}
\end{figure}

\begin{figure*}
\centering
\includegraphics[width=7.2
in]{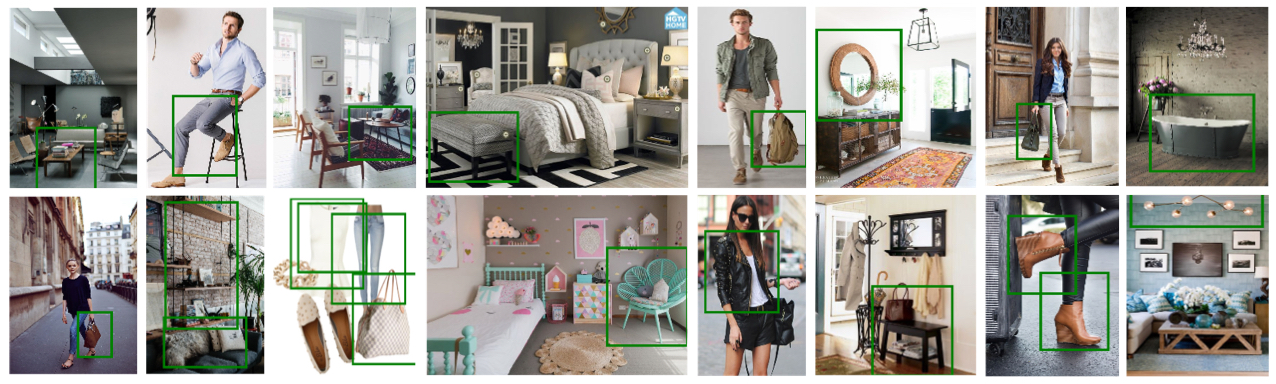}
\caption{Examples of detected visual objects from Pinterest's object detection pipeline. Detection of objects allows for more targeted visual features to be computed for Related Pins.}
\label{fig:object_detection}
\end{figure*}


In this subsection we present a qualitative analysis of using user curation signals to generate candidates for Related Pins. As we described previously, board co-occurrences for Pins is a strong signal of relevance and has been the foundation of candidate generation for our system. Figure~\ref{fig:occur_examples} illustrates that the relevance of the candidates grows gradually when the number of co-occurrences with the query Pin increases. 

We also show the percentage of image pairs having different number of board co-occurrences in Figure~\ref{fig:cooccurrence}. Note that the majority of the image pairs (around 80\%) only co-occur once on the same board, which suggests that ranking based on content features as the next step is important for finding high-quality recommendations. On the other hand, there are only a handful of image pairs which co-occur many times on the same boards.

\subsection{Ranking with Visual Features}

We found that one of the most important features for re-ranking the Pin candidates generated through board co-occurrence is visual similarity. We validated this by setting up a series of A/B experiments, where we selected five million popular Pins on Pinterest as queries, and re-ranked their recommendations using different sets of features. The control group re-ranked Related Pins using a linear model with a set of standard features: text annotation similarity, topic vector similarity, category vector similarity, as well as query-independent features. The treatment group re-ranked using fine-tuned VGG \textit{fc6} and \textit{fc8} visual similarity features along with indicator variables (in addition to the features used in control).

Across the 5M query Pins, the treatment saw a 3.2\% increase in save/clickthrough rate\footnote{This metric was measured across a 14 day period in Sep. 2015.}. After expanding the treatment to 100M query Pins, we observed a net gain of 4.0\% in propensity to engage with Related Pins, and subsequently launched this model into production. Similar experiments with a fine-tuned AlexNet model yielded worse results (only 0.8\% engagement gain).

When broken down by category, we noted that the engagement gain was stronger in predominantly visual categories, such as art (8.8\%), tattoos (8.0\%), illustrations (7.9\%), and design (7.7\%), and lower in categories which primarily rely on text, such as quotes (2.0\%) and fitness planning (0.2\%). Given the difference in performance among categories, we performed a follow-up experiment where we introduced a cross feature between the category vector of the query and the scalar \textit{fc6} visual similarity feature (between the query and candidate). This introduces 32 new features to the model, one for each of our site-wide categories (these features are sparse, since the Pinterest category vector thresholds most values to zero). The result from this was a further 1.2\% engagement increase in addition to the gains from the initial visual re-ranking model. Further work into component and cross product features is of interest to us, as they are essentially free to compute at rank-time, since the raw feature data is already stored.


\subsection{Ranking with Detected Objects}

\begin{figure}
\centering \includegraphics[width=2.2 in]{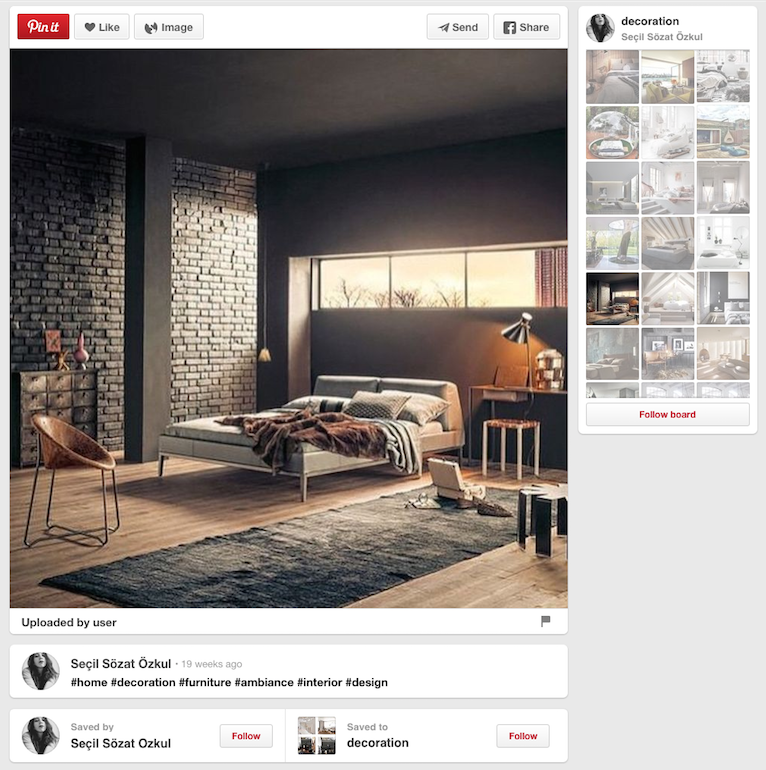}
\caption{Instead of generating recommendations for this entire interior design Pin, we may also want recommendations for the individual objects seen in the image, such as the chair on the left.}
\label{fig:visualobject}
\end{figure}

Users are sometimes interested in the \textit{objects} in the Pin's image, instead of the full image (as shown in Figure~\ref{fig:visualobject}).  We therefore speculate that object detection, when feasible, should improve relevance targeting. After applying non-maximum suppression (NMS) to the proposals generated by our fine-tuned Faster R-CNN module mentioned in section \uppercase\expandafter{\romannumeral4\relax}, we considered query Pins where the largest proposal occupies at least 25\% of the Pin's image, or if the proposal is smaller, it passes a confidence threshold of 0.9 in Faster R-CNN. We categorize these images as containing a dominant visual object, and using the best-performing fine-tuned VGG re-ranking variant from the previous section as our control, we experimented with the following treatments: 

\begin{itemize}
\item \textit{Variant A}: if a dominant visual object is detected in the query Pin, we compute visual features (VGG ) on just that object.
\item \textit{Variant B}: same as \textit{variant A}, but we also hand-tune the ranking model by increasing the weight given to visual similarity by a factor of 5. The intuition behind this variant is that when a dominant visual object is present, visual similarity becomes more important for recommendation quality.
\item \textit{Variant C}: if a dominant visual object is detected in the query Pin, we still use the features from the entire image (as the control does), but increase the weight given to visual similarity by a factor of 5, as in \textit{variant B}. In this variant, we assume that the presence of detected visual objects such as bags or shoes indicates that visual similarity is more important for this query. 
\end{itemize}

\begin{table}[h]
\centering
\begin{tabular}{  l  |  l |  c }

   \textbf{Features}  &  \textbf{Queries}  & \textbf{Engagement}    \\
\hline

FT-VGG (control)	 &  - & - \\
\hline
FT-VGG + category cross features	 & 5M & +1.2\% \\
\hline
FT-VGG + object detection \textit{variant A}	 & 315k fashion  & +0.5\%  \\
FT-VGG + object detection \textit{variant B}	 & 315k fashion & +0.9\%  \\
FT-VGG + object detection \textit{variant C}	 & 315k fashion & +4.9\%  \\

\end{tabular}
\caption{Results when using cross features and object detection, measured over a 7 day period in Oct. 2015}
\label{tbl:visualsearchrel}
\end{table}

Results for these variants are listed in Table \ref{tbl:visualsearchrel}. Variants A and B of the object detection experiments suggest that the tight bounding boxes from our object detection module do not provide enough context for our CNN models, but Variant C, which results in an additional 4.9\% engagement gain over the VGG similarity feature control, demonstrates that the presence of visual objects indicates that visual similarity should be weighed more heavily. Based on these results, our future focus is scaling up the number of object categories we can detect, and tuning the weight given to visual similarity in Variant B and C. 

\section{Conclusion and Future works}
The Related Pins system described in this work has improved user engagement metric and traffic on Pin recommendations by more than 50\% from May 2015 to November 2015. This demonstrates that signals derived from user curation and the activity of users organizing content contain rich information about the images and are very effective when used in conjunction with collaborative filtering. We also demonstrate that visual features such as representations learned from CNNs or presence of detected visual objects can be used in the learning-to-rank framework to improve item-to-item recommendation systems. One important component not discussed in this work is our use of user signal in the form of \textit{Navboost}, which also uses a model based on COEC (extended to actions beyond clicks) to re-rank content based on user engagement. Our future work includes exploring a richer set of features (e.g. sparse features, dense features, cross-product features, more object categories) and real-time recommendations (enabling re-ranking based on locale, current search query, and other forms of personalization).

\section{Acknowledgments}
We would like to thank our colleagues on the Visual Discovery and Recommendations teams at Pinterest, in particular Dmitry Chechik, Yunsong Guo, and many others. We'd also like to acknowledge Jeff Donahue and Trevor Darrell from Berkeley Vision and Learning Center (BVLC) for their collaboration with Pinterest and their work on Caffe.





\bibliographystyle{abbrv}

%





\end{document}